\RequirePackage{fix-cm}
\documentclass[twocolumn]{svjour3}          
\smartqed  
\usepackage{graphicx}
\usepackage{amsmath,amsfonts,amssymb}
\usepackage{graphicx}
\usepackage{setspace}
\usepackage{mathrsfs}
\usepackage{multirow}
\begin{document}

\title{Can We Boost the Power of the Viola$-$Jones Face Detector Using Pre-processing? An Empirical Study
}


\titlerunning{Toward Boosting the Power of the Viola$-$Jones Face Detector}        

\author{Mahmoud Afifi         \and
        Marwa Nasser \and 
        Mostafa Korashy \and
        Katherine Rohde \and
        Aly Abdelrahim 
}

\authorrunning{Mahmoud Afifi et al.} 

\institute{Mahmoud Afifi \at
Information Technology Department, Assiut University, Egypt\\
\email{mafifi@eecs.yorku.ca}\\
\emph{Present address:}  Department of Electrical Engineering and Computer Science, Lassonde School of Engineering, York University, Canada
\and
Marwa Nasser \at
Information Technology Department,Faculty of Computers and Information, Assiut University, Egypt
\and
Mostafa Korashy \at
Information Technology Department,Faculty of Computers and Information, Assiut University, Egypt
          \and
          Katherine Rohde
          \at
          Radiology Department, University of Pittsburgh, USA
          \and
          Aly Abdelrahim \at
          Computer Science Department, Faculty of Computers and Information, Assiut University, Egypt            
}

\date{}

\maketitle

\begin{abstract}
The Viola-Jones face detection algorithm was (and still is) a quite popular face detector. In spite of the numerous face detection techniques that have been recently presented, there are many research works that are still based on the Viola-Jones algorithm because of its simplicity. In this paper, we study the influence of a set of blind pre-processing methods on the face detection rate using the Viola-Jones algorithm. We focus on two aspects of improvement, specifically badly illuminated faces and blurred faces. Many methods for lighting invariant and deblurring are used in order to improve the detection accuracy. We want to avoid using blind pre-processing methods that may obstruct the face detector. To that end, we perform two sets of experiments. The first set is performed to avoid any blind pre-processing method that may hurt the face detector. The second set is performed to study the effect of the selected pre-processing methods on images that suffer from hard conditions. We present two manners of applying the pre-processing method to the image prior to being used by the Viola-Jones face detector. Four different datasets are used to draw a coherent conclusion about the potential improvement caused by using prior enhanced images. The results demonstrate that some of the pre-processing methods may hurt the accuracy of Viola-Jones face detection algorithm. However, other pre-processing methods have an evident positive impact on the accuracy of the face detector. Overall, we recommend three simple and fast blind photometric normalization methods as a pre-processing step in order to improve the accuracy of the pre-trained Viola-Jones face detector.
\end{abstract}

\section{Introduction}
\label{sect:intro}  
Although many face detection techniques have been presented in the literature, face detection is still deemed one of the most challenging tasks in the field of computer vision. Numerous applications are based on detecting human faces, such as facial recognition, social media, gaming, marketing, augmented reality, and smart surveillance systems. In the literature, there are a few number of studies that investigate the impact of illumination pre-processing methods on the face recognition process \cite{re1,re2,re3}. 
\\
In the comparative study \cite{re1}, several illumination compensation and normalization techniques were applied to three datasets in order to study the influence of such techniques on the face recognition process. The self-quotient image (SQI) \cite{SQI} and the modified local binary pattern (mLBP) \cite{mlpb} obtain the best recognition results using different eigenspace-based face recognition approaches. Three different datasets were used in the study proposed by R. Gopalan and D. Jacobs \cite{re2} who suggest that the gradient direction of the face image \cite{GradDir} and SQI improves the recognition rate under different lighting conditions.
\\
In the study \cite{re3}, the effect of different illumination insensitive techniques on face recognition was studied using four different datasets. The study recommended that intensity transformation methods, such as histogram equalization and logarithmic transform, can be used to improve the accuracy of recognition. The study argued that those methods may improve the face detection process as well. Although the impact of poor illumination conditions on face detection accuracy is well-known, there is no dedicated study, to the best of our knowledge, that investigates through experimentation the impact of the illumination pre-processing methods on the face detection process. Additionally, blurry faces may evade the face detector; thus, deblurring and sharpness enhancement may improve the detection accuracy. Fig. 1 
 shows the performance of the Viola-Jones algorithm \cite{Viola} using the Yale Face Database A \cite{Yale} which has 99.39\% of their images that are detected by the Viola-Jones face detector; however, the performance is affected by the synthetic blur which was added using $n\times n$ average blur kernel, where $n\in\{3,11,19,27,35,43\}$.
\\
According to S. Zafeiriou et al. \cite{survey}, the face detection techniques can be categorized into three main categories: (1) boosting-based algorithms that use a combination of multiple weak classifiers. (2) Algorithms that are based on deep learning, where deep convolutional neural networks are utilized in order to detect faces; these algorithms usually obtain an impressive accuracy and considered the state-of-art techniques. (3) Face detection based on deformable model.
\\
The Viola-Jones face detection algorithm, which belongs to the first category, is one of the most widely used face detectors because of its efficiency, effectiveness, and simplicity. The Viola-Jones object detection is considered the first strong framework that achieves high detection rates in real-time usage. Furthermore, the Viola-Jones face detector is well trained and tested, and it is robust against harsh conditions. Although the Viola-Jones algorithm is considered a relatively old approach for face detection, it is still under improvement and used in many applications. Recently, P. Irgens et al. \cite{hardware} presented a complete system level hardware design of the Viola-Jones algorithm. V. Mutneja et al. \cite{gpu} presented a GPU-based modified Viola-Jones algorithm in order to accelerate the training process of the algorithm. S. Tikoo \cite{cnn} presented a framework for face detection and recognition by combining both the Viola- Jones algorithm and back propagation neural network. S. El Kaddouhi \cite{eye} have used a two-stages Viola-Jones-based face detector in order to detect eye region on face images.
\\
The Viola-Jones face detector consists of three main stages. Feature extraction, boosting, and cascading. First of all, Haar-like features are calculated. However, these simple features are considered weak, because many objects may match the pattern. AdaBoost algorithm \cite{ada} is used to combine the most strong classifiers among those weak ones by a supervised learning within the training stage. A Cascade of classifiers is used by grouping the Haar features into different levels to form the final classifier.
\\
In this paper, we aim to improve the detection rate of the Viola-Jones algorithm by manipulating the image prior to being used in the face detection stage rather than using a complicated detection approach. To that end, we present an empirical study on the impact of a set of image pre-processing techniques on the face detection process. We use two main categories of pre-processing methods. The first category aims to produce a lighting invariant face image to be used by the face detector, such as intensity transformation, retinex, gradient normalization, and homomorphic filtering. The second category's target is to deblur the original face image using sharpness and deblurring techniques. Many techniques are applied in the pre-processing stage to study the effect of each one on the face detection accuracy obtained by the Viola-Jones face detector. Our study is targeted towards general images that may or may not include bad illumination conditions or blurry faces. We could deal with the problem of low-light and blurry face images by re-training the Cascade using, for example, blurry face images. However, the trained Cascade is then directed only for this kind of face image. Consequently, that leads to the need for a robust image sharpness assessment technique that works properly with blurry face images to tell which Cascade is more appropriate, i.e., the regular Cascade or the classifier that was trained by blurry face images. The image assessment techniques usually deal with the whole image, which is considered a misleader in our case. Although the two images in Fig. 2 are for the same face image, represented by sharp face pixels in the first image (a) and blurry pixels in the second image (b), the sharpness measure $h$ by Rania Hassen et al. \cite{imageassessment} is similar for each of them (0.954 and 0.9547). In most blurry face images, the camera focuses on the background instead of the face; that makes the sharpness measure similar in the case of small face regions, relative to the background region. Thus, the image assessment approaches seem to be inapplicable in many scenarios of blurry face images. The same is true in poorly illuminated face images. For that reason, we use the pre-processing methods in a blind manner in order to boost the true detection rate of the Viola-Jones algorithm using a pre-trained Cascade classifier.
\\
In order to draw a coherent conclusion, we perform two set of experiments. In the first set of experiments, we use three general benchmark datasets that are considered simple datasets for face detection. Additionally, we discard any non-frontal face images in the datasets. Frontal and simple face images were chosen because we want to exclude any pre-processing methods that may hurt the performance of the Viola-Jones algorithm. Thus, we use simple face images that are expected to produce a good true detection rate using the pre-trained Viola-Jones algorithm without any pre-processing. Consequently, we can exclude any pre-processing method that reduces the true detection rate.
\\
In the second set, we use a dataset that suffers from harsh lighting conditions and contains many blurry face images. In this set of experiments, we are not constrained with frontal images; we also deal with near-frontal face images. The second set of experiments helps us to understand how the pre-processing methods can improve the Viola-Jones algorithm's performance on low-light and/or blurry face images.
\\
The rest of this paper is organized as follows: Section 2, describes the pre-processing methods to improve face detection results based on
correcting input images. Section 3 shows the datasets that are used. Experimental results are presented in Section 4, and the paper is concluded in Section 5.

\begin{figure}[t]
\centering
\label{fig1}
\includegraphics[width=\linewidth]{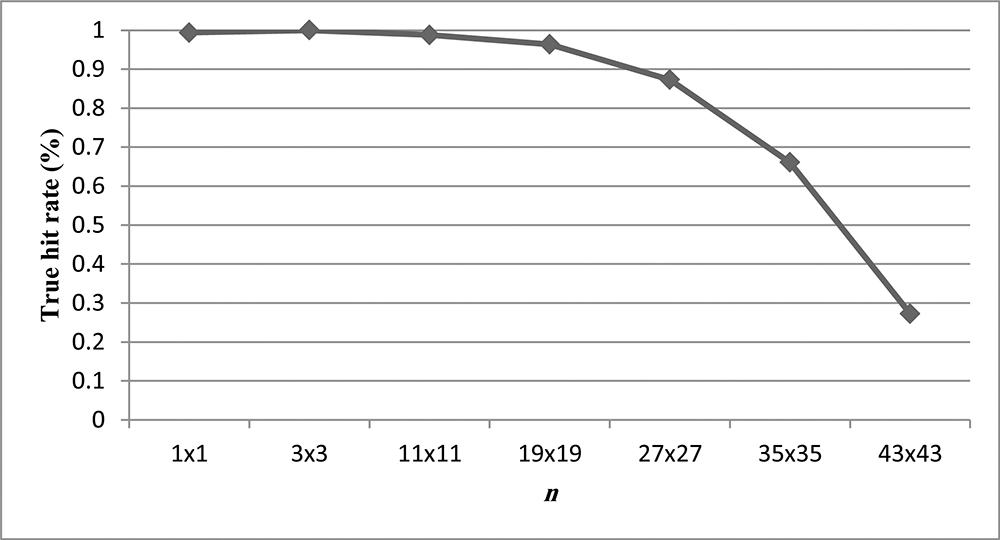}
\caption{The true hit rate of the Viola-Jones algorithm \cite{Viola} using 7 versions of the Yale Face Database A \cite{Yale}. Each version contains blurred images using $n\times n$ average blur kernel.}
\end{figure}

\begin{figure*}[t]
\centering
\label{fig5}
\includegraphics[width=0.7\linewidth]{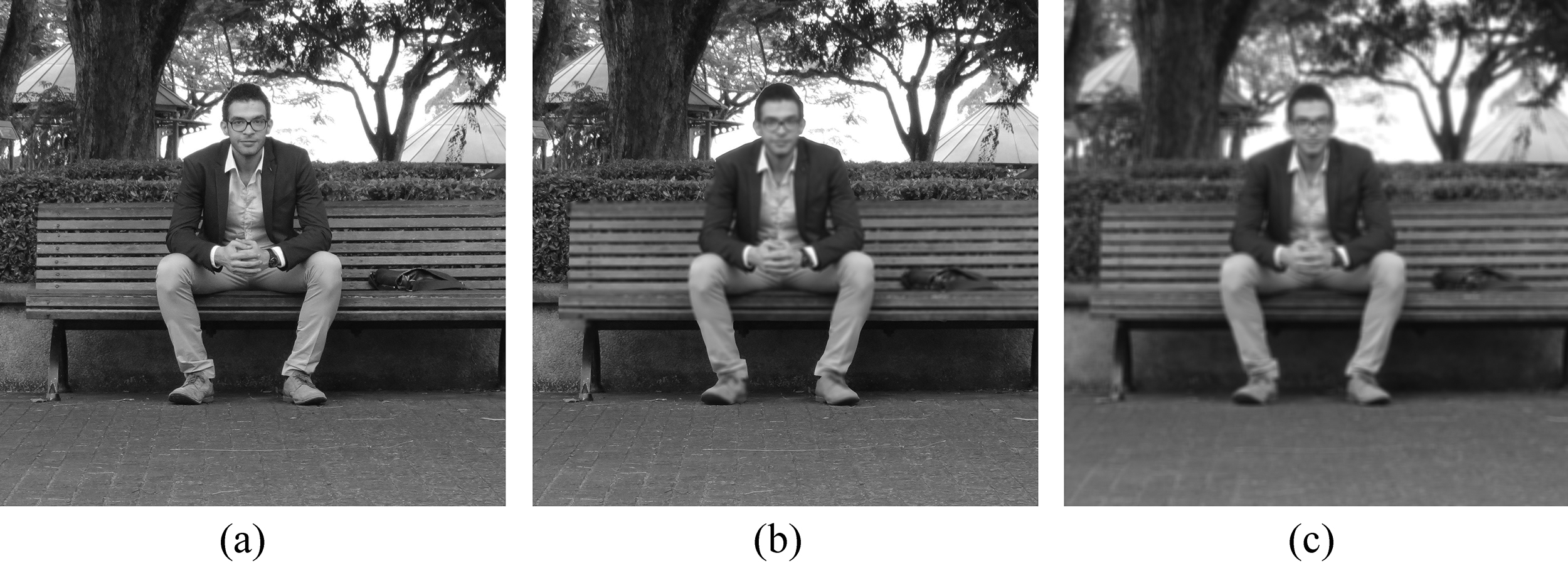}
\caption{The sharpness measure $h$ \cite{imageassessment} of face images. (a) Original image. (b) Blurry face region. (c) Completely blurred image. The sharpness measure $h=0.9540$ for (a), $h=0.9547$ for (b), and $h= 0.7096$ for (c).}
\end{figure*}

\section{Pre-processing methods}
In order to test the rate of possible improvement caused by using the pre-processing methods, we test two manners of applying the pre-processing methods. In the first manner, the pre-processing methods are simply applied prior to the Viola-Jones face detector, which shows the real effect of the pre-processing method. The second manner adjusts the generated image of the pre-processing method. The intensity of each pixel is mapped in order to adjust the image intensity of bright or dark generated images. Therefore, there is a new pre-processing step applied after producing the image generated by the pre-processing method. We have tested both histogram equalization and contrast stretching in order to adjust the generated image, and the last one gives visually better results; consequently, we have used contrast stretching in the image adjustment process. The face detector is then applied to the adjusted image (see Fig. 3).
\\
The pre-processing methods are categorized into two main groups. The first group is directed to get a lighting invariant version of the given face image. The second group aims to diminish as much as possible blurring artifacts from the given image. We used a diverse of techniques starting from simple methods to sophisticated ones. In the first group, we use the single-scale retinex (SSR) \cite{SSR}, the multi-scale retinex (MSR) \cite{MSR}, the adaptive single scale retinex (ASSR) \cite{ASSR}, the non-local-means-based normalization technique (NMBN) \cite{ANL}, the homomorphic filtering based normalization (HOMO), the Discrete cosine transform (DCT) \cite{DCT}, the gradient normalization \cite{GRF}, the large- and small-scale features normalization technique \cite{LSSF}, the predefined distribution fitting of histogram (PDF) \cite{FITT}, and the non point light and error quotient image (NPLE-QI) \cite{NPLEQ}. In the second group, we utilize standard sharpness, the Wiener filtering \cite{Wiener}, the blind deconvolution algorithm (BDA) \cite{BDA}, the blind motion deblurring (BMD) \cite{MotionBlur}, and the general framework for image restoration (GFIR) \cite{GeneralFramework} are used. Fig. 4 shows the results of the pre-processing methods. In the following paragraph we briefly describe each of the ten lighting enhancement methods that have been used in this study. \\

\begin{figure}[t]
\centering
\label{fig6}
\includegraphics[width=\linewidth]{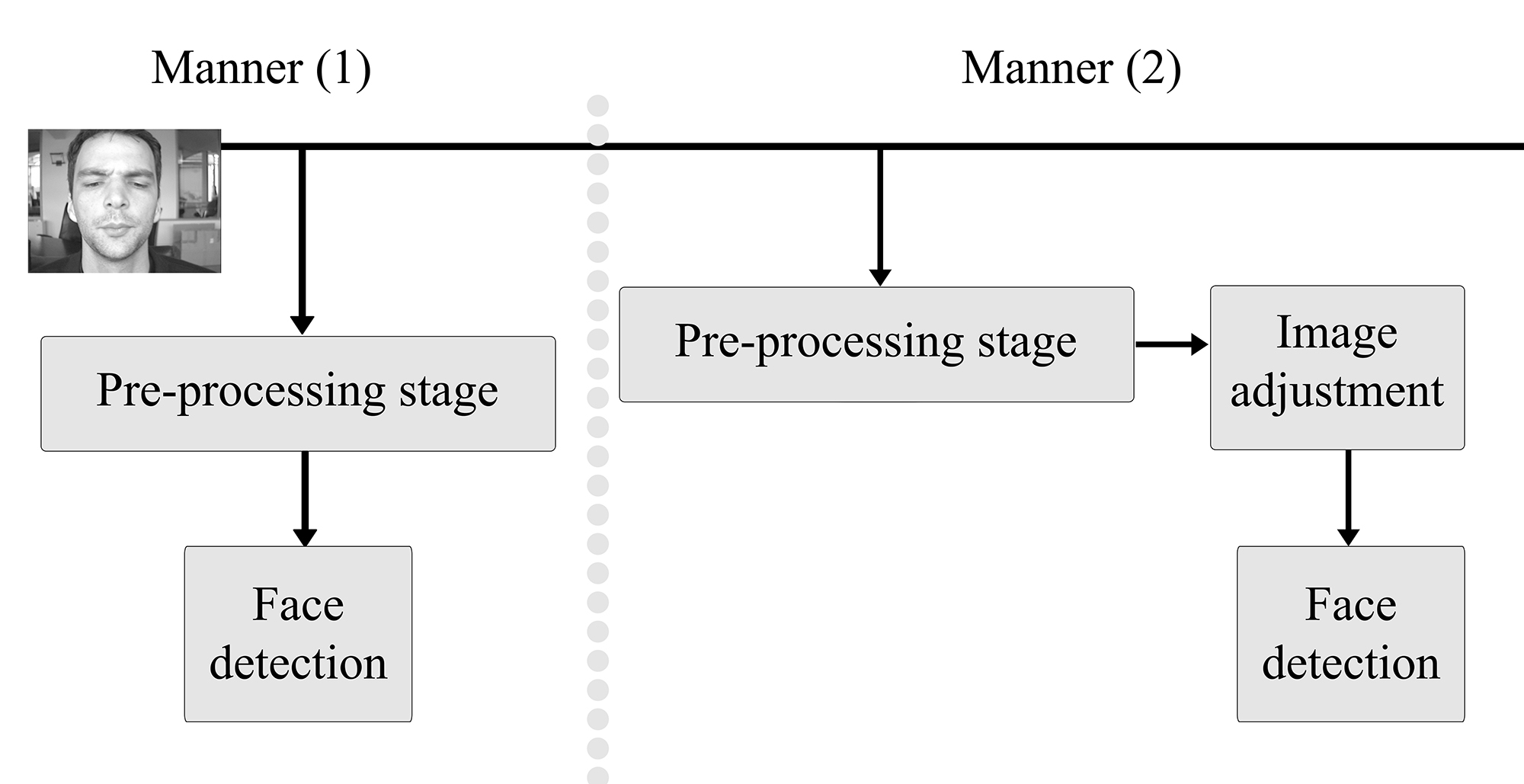}
\caption{The two manners of applying the pre-processing methods prior the face detection stage.}
\end{figure}

\begin{figure}[t]
\centering
\label{sampleImages}
\includegraphics[width=\linewidth]{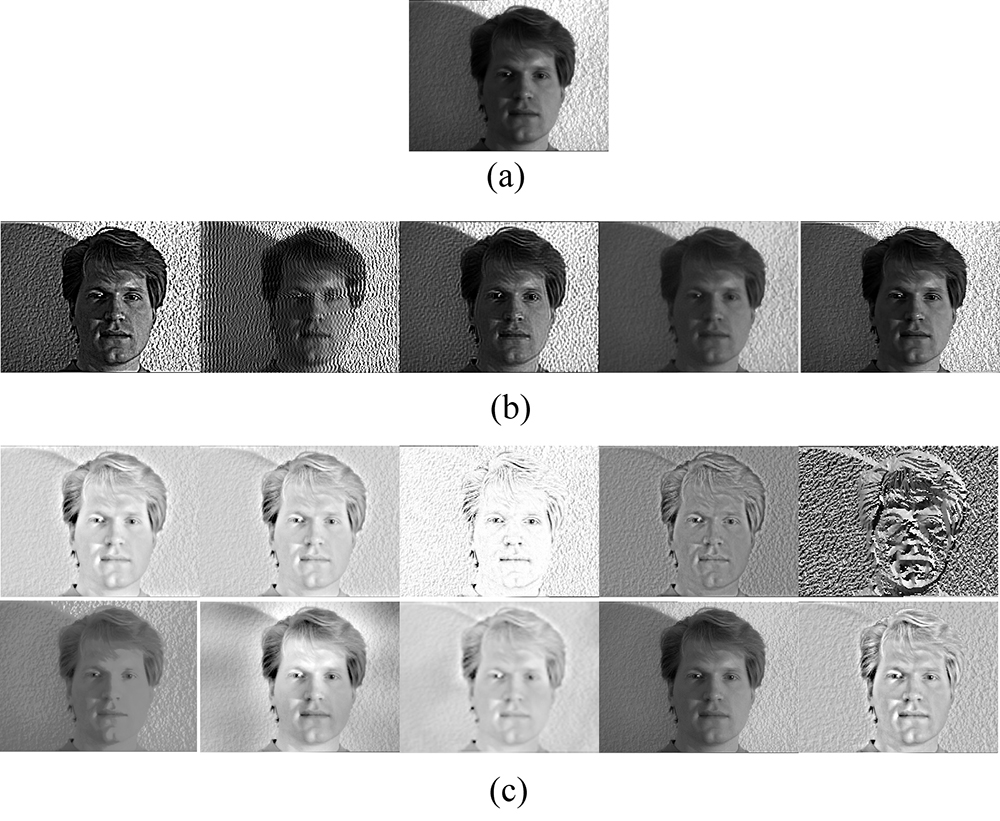}
\caption{The results of the pre-processing methods prior applying the Viola-Jones face detector. The original image, which is extracted from the Yale Face Database A [9], is shown in (a). Followed by the results of the deblurring methods in (b). The deblurring methods, from left to right, are Sharpness, Wiener, BDA, GFIR, and BMD. The results of the lighting enhancement methods are shown in (c). The methods, from left to right, are SSR, MSR, ASSR, LSSF, Gradient normalization, PDF, DCT, NPLE-QI, HOMO, and NMBN. }
\end{figure}

\textbf{SSR: }In the single-scale retinex (SSR) \cite{SSR}, the illumination is estimated using a smoothed version of the image obtained by using a Gaussian linear low-pass filter (LPF). The log of estimated global illumination is then subtracted from the log of the image.
\\
\textbf{MSR:} The multi-scale retinex (MSR) \cite{MSR} is considered an improvement of the SSR by using multiple lighting invariant SSR images.
\\
\textbf{ASSR:} Instead of using a smoothed image that is generated by a static low-pass filter, adaptive single scale retinex (ASSR) \cite{ASSR} is based on an adaptive smoothing manner to obtain the illumination of the image that is then divided by the estimated illumination as in the SSR.
\\
\textbf{NMBN:} An enhancement for the well known Non-Local Means Algorithm (NL Means) is used \cite{ANL}. The weighting function $w_{(z,x)}$ of NL means is defined as follows:
\begin{equation}
w_{(z,x)}=\frac{1}{Z(z)} e\tfrac{G_\sigma||I_n(\sigma_x)-I_n(\sigma_z)||_{2}^{2}}{h^2},
\end{equation} 
\begin{equation}
Z(z)=\sum_{x\in I_n(x)}e\tfrac{G_\sigma||I_n(\sigma_x)-I_n(\sigma_z)||_{2}^{2}}{h^2}
\end{equation}
\\
In the previous equation, $h$ stands for the parameter that controls the decay of the exponential function, by using $h$ as a function of local contrast instead of being a fixed predetermined value. This effect has dramatically enhanced the results of using the original NL means algorithm.
\\
\textbf{HOMO:} Homomorphic filtering based normalization (HOMO) is performed by transforming the image into the frequency domain to reduce the low-frequencies and emphasize the high-frequencies that contain the details of the image \cite{HOMO}.
\\
\textbf{DCT:}  Discrete cosine transform (DCT) is used in the normalization \cite{DCT} by trimming low frequencies of the image in the frequency domain.
\\
\textbf{Gradient:} The gradient normalization \cite{GRF} uses the orientation of the gradients of the image to generate a lighting invariant face image. However, it generates a distorted image from the aspect of the visual human perception.
\\
\textbf{LSSF:} By composing the image into large and small scale features and applying an illumination correction to the set of large-scale features and only small illumination corrections are made to the small-scale features, the resulting image is produced by combining the corrected large-scale and small-scale features.
\\
\textbf{PDF:}  The authors of \cite{FITT} investigated replacing the distribution of the given image by other arbitrary distributions such as normal, lognormal and exponential distributions. Enhanced and similar results are obtained using other distributions instead of using the normal distribution; however, many efforts are required in selecting other distributions parameters.
\\
\textbf{NPLE-QI:} The authors of \cite{NPLEQ} have proposed an extension for quotient image-based illumination normalization by considering cast shadows in the process of extracting large-scale illumination invariant features. The technique has proven promising results under difficult illuminations with cast shadows.
 \\
 In the following paragraph, we will briefly describe each of the five deblurring methods used in this study.
 \\
 \textbf{Sharpness:} By adding the Laplacian of the image $\bigtriangledown ^{2} f$ multiplied by a center coefficient indicator $c$ to the original image $f$, the sharp image $\widehat{f}$ is generated by:
 \begin{equation}
 \widehat{f}_{(x,y)}=f_{(x,y)}+c[\bigtriangledown ^{2}f_{(x,y)}]
  \end{equation}
 \textbf{Wiener:} By convolving the degraded image (i.e. blurred image) $g$ with the degradation filter, namely the Wiener \cite{Wiener} filter given by:
 \begin{equation}
  W_{(x,y)}=\frac{H^*_{(x,y)}P_{s(x,y)}}{|H_{(x,y)}|^2P_{s(x,y)}+P_{n(x,y)}}
  \end{equation}
 \textbf{BDA:}  The blind deconvolution algorithm (BDA) \cite{BDA} is performed by estimating the PSF. The recovered PSF is obtained using the maximum likelihood algorithm. The restored image is obtained by applying the deconvolution process with the recovered PSF to the degraded image.
 \\
 \textbf{BMD:} Blind motion deblurring (BMD) \cite{MotionBlur} is performed by estimating the blur filter. The restored image is obtained by applying the deconvolution process with the recovered blur filter to the degraded image. The recovered filter is estimated by solving an optimization problem to taking into account salient edges and low rank prior.
 \\
 \textbf{GFIR:} In the general framework for image restoration (GFIR) \cite{GeneralFramework}, a large-scale framework for kernel similarity-based image restoration has been presented. The technique consists of inner and outer loops. In each iteration in the outer loop, the similarity weights are recalculated using the previous estimated values; while in the inner loop, the updated objective function is minimized using inner conjugate gradient iterations.

\section{Datasets}
As aforementioned, we perform two different set of experiments. In the first set we use three face datasets in order to discard any pre-processing method that may hurt the performance of the Viola-Jones algorithm. In the second set, we use a dataset that contains many blurry, occluded, and low-light face images in order to draw our conclusions about the impact of the pre-processing methods on the accuracy of the Viola-Jones face detection algorithm. Fig. 5 shows samples of each dataset that have been used in this work. The first set of experiments uses the following datasets: 1- The ORL database of faces \cite{ORL}, 2- the MIT-CBCL face recognition database \cite{MIT}, and 3- the BioID face database \cite{BioID}. The second set of experiments uses the Specs on Face (SoF) dataset \cite{SoF}.

\subsection{The ORL database of faces}
The ORL database of faces consists of 400 (92$\times$112 pixels) frontal face images for 40 subjects. Many subjects were captured under different lighting conditions with several facial expressions. Unfortunately, face annotations are not supported.

\subsection{The MIT-CBCL face recognition database}
The MIT-CBCL face recognition database contains 2,000 (115 $\times$115 pixels) frontal face images that were generated by projecting 3D synthetic models of 10 different subjects to 2D images. As aforementioned, this work focuses on frontal face images; thereby non-frontal images were excluded so that the number of face images is reduced to be 772. The face regions are not supported as well.

\subsection{The BioID face database}
The BioID face database contains frontal and non-frontal images for 23 different subject. The dataset comprises 3,043 (384$\times$286 pixels) face images that is reduced to 1,521 frontal face images after removing non-frontal face images. The dataset comes with a ground-truth landmarks associated with each image.

\subsection{The Specs on Face dataset}
The SoF dataset \cite{SoF} contains frontal and non-frontal for 112 persons (66 males and 46 females) with different facial expressions under harsh illumination environments. The dataset comprises 2,662 (640$\times$ 480 pixels) face images. All images contains people who wear eyeglasses as a common facial occlusion in the dataset. Besides the original face images, the dataset contains three groups of synthetic images: noisy, blurry, and posterized face images with three levels of difficulty (easy, medium, and hard). Since we focus on the problem of badly illuminated and blurred faces, we deal only with the original and the sets of blurry face images (easy, medium, and hard). The total number of face images we use from the SoF dataset is 10,648 images.

\begin{figure}[t]
\label{sample_iamges_dataset}
\centering
\label{fig7}
\includegraphics[width=\linewidth]{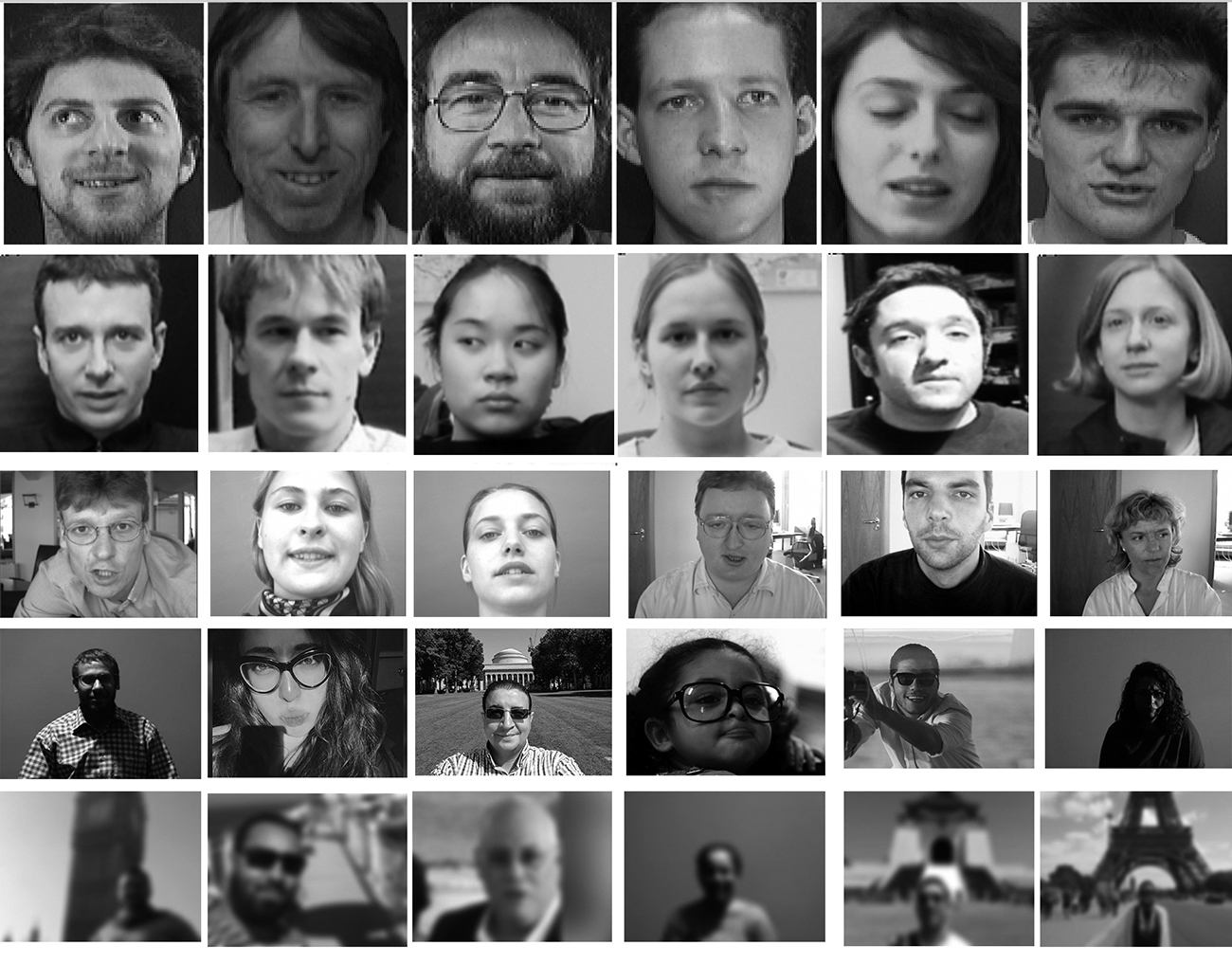}
\caption{Sample images of the datasets that have been used in this study. The first three rows show samples from the datasets, namely ORL, MIT-CBCL, and BioID datasets,that have been used in the first set of experiments. The last two rows show samples from the SoF dataset that has been used in the second set of experiments.}
\end{figure}

\section{Experimental results}
We adopted the Matlab pre-trained Cascade classifier of the Viola-Jones face detector. Since the first three datasets, mentioned in Section 3, have not sup- ported face annotations, handcrafted face labels were determined carefully by three different persons. We have used the face annotations provided by the SoF dataset. According to the ground-truth face regions, the detected regions with an Intersection-Over-Union (IoU) score that exceeds 50\% were accepted as a true-positive detection. Otherwise, the detected regions is considered false-positive detection. The IoU is calculated by
\begin{align}
\begin{split}
score=\frac{area(A\cap B)}{area(A\cup B)},
\end{split}
\end{align}
\noindent
where $A$ is the ground-truth face ROI and $B$ is the detected face rectangle. 
\\
The Gaussian filter that was used in the MSR consisted of 7 rows and 15 columns using 21 iterations. The ASSR was carried out using 15 iterative convolutions. The low-frequency which is corresponded to the DCT coefficients was 20. In the homomorphic filtering, we have used 2 as the ratio that high frequency values are boosted relative to the low frequencies. The cut-off frequency of the HOMO filter was 0.25. In the PDF method, we have used a normal distribution with mean value $m=0$ and standard deviation value $\sigma=1$. In the NPLE-QI, the number of illumination basis was 20 using a weight for error basis equals to 0.1 and 0.03 as the parameter of fitting error. We have used the motion filter as the PSF for Wiener filtering. The linear motion of a camera, in pixels, was the size of each image divided by 40. The angle of the motion filter equals to the size of the image divided by 30 in the counter-clockwise direction. In the BDA, we have used the 5$\times$5 rotationally symmetric Gaussian low-pass filter using standard deviation equals to 7. In the BMD, a 19$\times$19 kernel was used with the same parameters specified in the paper \cite{MotionBlur}. In the GFIR, we adopted the Gaussian blurring scenario with 11 $\times$ 11 kernel and standard deviation $\sigma=0.9$. All above parameters were experimentally determined.
\\
We conducted two sets of experiments. In the first set of experiments, all pre-processing methods were used using face images consisting of mostly frontal faces in controlled environments without any harsh conditions involved. After analyzing the results of this dummy test, we found that some pre-processing methods may hurt the accuracy of the face detector. Consequently, we excluded these methods from the second set of experiments which is performed using a dataset of face images with many different hard conditions. The goal of the second set of experiments is to determine the potential for improvement using the pre-processing methods when the face images suffer from bad conditions, i.e. blur or bad lighting conditions. In this section, we report the results of both sets of experiments in order to draw the final conclusion in Section 5.
\subsection{First set of experiments}
All of the aforementioned pre-processing methods were applied as blind methods to 2,693 face images, collected from the first three datasets discussed previously, followed by applying the Viola-Jones face detector.
Fig. 6 shows the performance of the Viola-Jones algorithm without using any pre-processing step prior to the face detection process. Fig. 7 shows the over- all recall, precision, and TP rate of the pre-processing methods by performing each discussed manner using the 2,693 face images. The precision scores obtained by the HOMO and PDF using the two manners outperform the precision score achieved without any pre-processing stage. NPLE-QI achieves a higher precision score using the second manner than the score obtained without any pre-processing methods. The BDA and BMD only improve slightly in regards to precision. Although the recall scores of the PDF and the GFIR methods using the two manners are considered high, these scores are below the original one by the Viola-Jones algorithm.
\\
From another viewpoint, the true positive rate (TP) of the Viola-Jones algorithm is improved using GFIR, HOMO, and PDF with both manners. Additionally, the NPLE-QI improves the TP using the second manner. The improvements obtained by using the GFIR, the PDF, the NPLE-QI, and the HOMO are very small; however, they, at least, do not hurt the Viola-Jones detector.
\\

The gradient normalization is considered the worst choice that drops down the recall to be 3.31\% and 3.11\%. The precision is also dropped down to be
85.57\% and 86.59\%. Finally, the TP rate goes down to be 3.30\% and 3.12\%
using the first and second manners, respectively.

\subsubsection{Analysis}
Table \ref{Table1_1} illustrates the TP, also known as positive hit rate, obtained by using the first manner of applying the pre-processing step. As shown most of the pre-processing methods hurt the Viola-Jones face detector except the GFIR, the HOMO, and the PDF, which increase the TP rate by 0.11\%, 0.21\%, and
0.41\%, respectively. However, we can see the fluctuation of the TP rate with the datasets. For example, the GFIR increases the TP rate by approximately
0.25\% using the ORL; nevertheless, the true hit rate goes down by 0.38\% using 
the MIT-CBCL dataset. Again, the gradient normalization is considered the worst choice, where it decreases the TP rate by 89.12\%. Table 2 shows the FP rates that were caused by applying the pre-processing methods using the first manner. There is no clear pattern of the changes that occurred by the pre-processing methods; some methods increase the FP rate. Others decrease it. All methods increase or have no effect on the FP rate with the MIT-CBCL dataset and decrease it with the ORL dataset, except the GFIR which increases it by 2\%. Some methods decrease the FP rate with the BioID dataset, and others increase it. 
\begin{figure}[t]
\centering
\label{fig8}
\includegraphics[width=\linewidth]{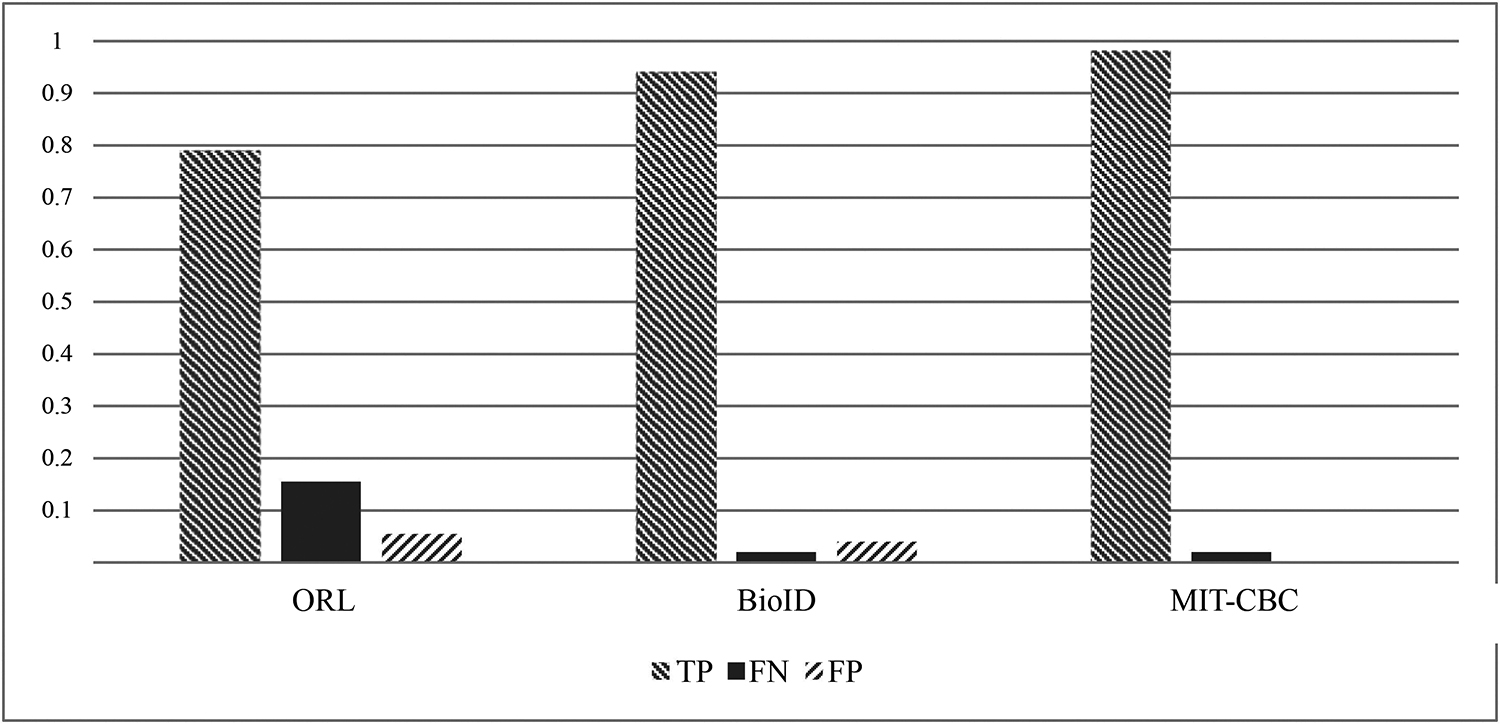}
\caption{Results obtained by using the Viola-Jones face detector without any pre-processing steps. TP refers to true positive rate, FP refers to false positive, and FN refers to false negative, i.e. undetected faces.}
\end{figure}
\noindent

\begin{figure*}[t]
\centering
\label{fig9}
\includegraphics[width=0.79\linewidth]{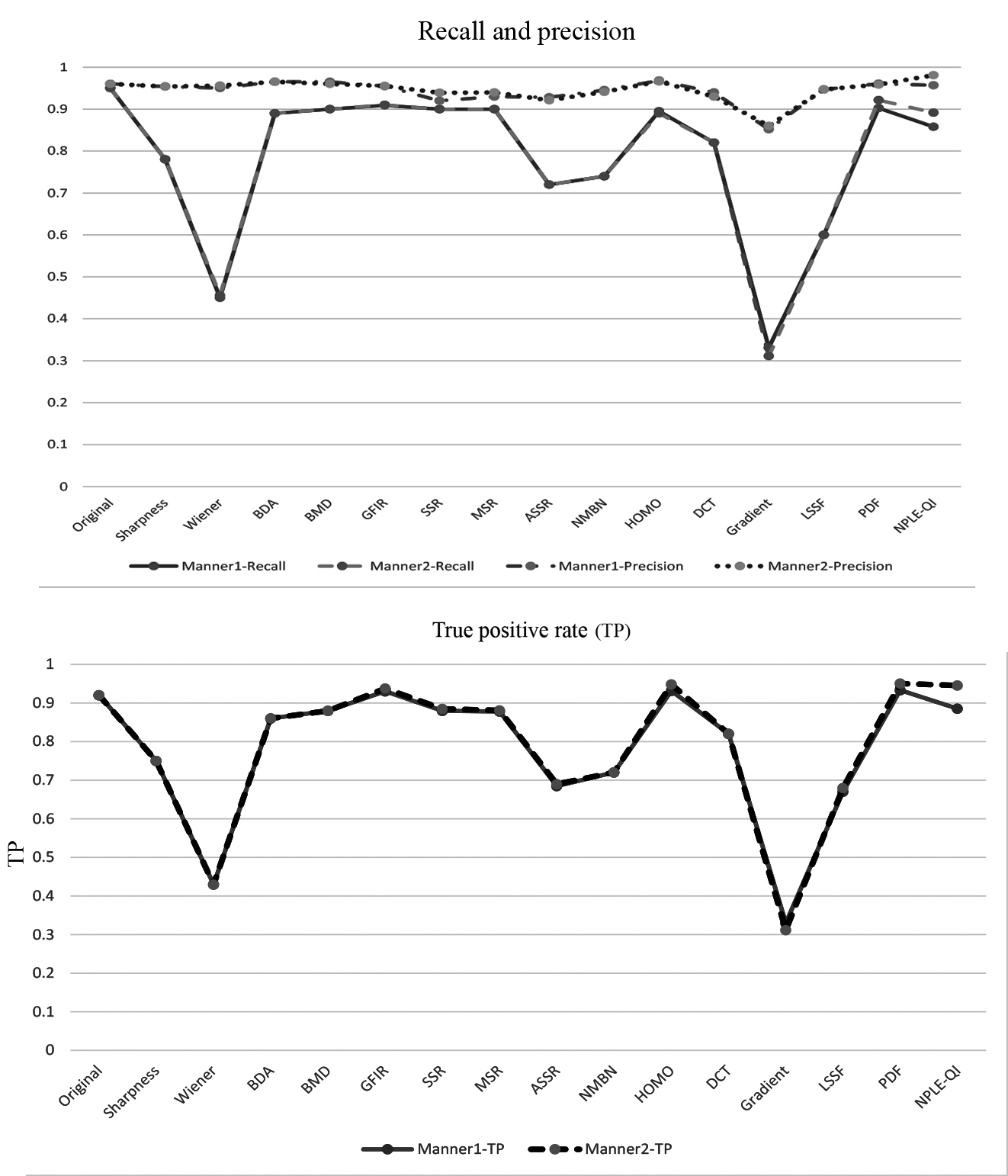}
\caption{The recall, precision, and true positive rate achieved by the pre-processing methods using the mentioned manners. The reported results were obtained using 2,693 face images from the different 3 datasets.}
\end{figure*}

\begin{table}[]
\centering
\caption{The effect of the pre-processing methods on the true positive rates (\%) using the manner 1.}
\label{Table1_1}
\begin{tabular}{l|l|l|l|l}
\cline{2-4}
                                              & \multicolumn{3}{c|}{\textbf{Datasets}}                    &                                     \\ \hline
\multicolumn{1}{|l|}{\textbf{Methods}} & ORL & MIT-CBCL & BioID &  \multicolumn{1}{l|}{\textbf{Total}} \\ \hline
\multicolumn{1}{|l|}{Sharpness}               & -24

         & -4.66

                      & -21.7

                                         &	
                                          \multicolumn{1}{l|}{-17.16
                                          }               \\ \hline
\multicolumn{1}{|l|}{Wiener}                  &  -15.75

          & -3.37

                       &  -80.6

                                           & \multicolumn{1}{l|}{-48.83}               \\ \hline
\multicolumn{1}{|l|}{BDA}                     &  -11.5

         &  -1.036

                     &  -7.56

                                        & \multicolumn{1}{l|}{-6.28
                                        }               \\ \hline
\multicolumn{1}{|l|}{GFIR}&+0.25&  -0.38&  +0.33&  \multicolumn{1}{l|}{+0.11}               \\ \hline

\multicolumn{1}{|l|}{BMD}                     &  -24.25

         & -3.11

                      &  +0.26

                                         & \multicolumn{1}{l|}{-4.34}               \\ \hline
\multicolumn{1}{|l|}{SSR}                     &    -12

       &   -1.55

                  &  -3.68

                           & \multicolumn{1}{l|}{-4.31
                           }               \\ \hline
\multicolumn{1}{|l|}{MSR}                     & -10.5
       &  -2.073

                  &  -4.08

                                    &
                                  \multicolumn{1}{l|}{-4.46
                                  }               \\ \hline
\multicolumn{1}{|l|}{ASSR}                    &  -24.5

         &  -8.55

                    &  -31.36
                        & \multicolumn{1}{l|}{-23.80
                        }               \\ \hline
\multicolumn{1}{|l|}{LSSF}                    & -33.75

          & -5.96

                       &  -49.05

                               & \multicolumn{1}{l|}{-34.42
                               }               \\ \hline
\multicolumn{1}{|l|}{Gradient}                    &     -78.5

    &  -85.1

               &   -93.95
                   & \multicolumn{1}{l|}{-89.12
                   }               \\ \hline
\multicolumn{1}{|l|}{PDF}                     & -0.75

           &  +0.26

                     &  +0.8

                              & \multicolumn{1}{l|}{+0.41

                              }               \\ \hline
\multicolumn{1}{|l|}{DCT}                & -32

          &   -16.84

                    &  -2.56

                            & \multicolumn{1}{l|}{-11.03
                            }               \\ \hline
\multicolumn{1}{|l|}{NPLE-QI}                    &  -12

         &   -6.99

                   &   -1.05
                          & \multicolumn{1}{l|}{-4.38
                          }               \\ \hline
\multicolumn{1}{|l|}{HOMO}                     & +0.25

          &  +0.39

                    &  +0.1
                            & \multicolumn{1}{l|}{+0.21
                            }               \\ \hline
\multicolumn{1}{|l|}{NMBN}                 &-28.5

           &  -6.35

                      &  -25.77

                              & \multicolumn{1}{l|}{-20.61
                              }               \\ \hline
\end{tabular}
\end{table}

\begin{table}[]
\centering
\caption{The effect of the pre-processing methods on the false positive rates (\%) using the manner 1.}
\label{Table1_2}
\begin{tabular}{l|l|l|l|l}
\cline{2-4}
                                              & \multicolumn{3}{c|}{\textbf{Datasets}}                    &                                     \\ \hline
\multicolumn{1}{|l|}{\textbf{Methods}} & ORL & MIT-CBCL & BioID &  \multicolumn{1}{l|}{\textbf{Total}} \\ \hline
\multicolumn{1}{|l|}{Sharpness}              & -5
       & +2.07

                    & -0.72

                                         & \multicolumn{1}{l|}{-0.56

                                         }               \\ \hline
\multicolumn{1}{|l|}{Wiener}                  &  -5

         & +2.72

                      &  -3.68

                                          & \multicolumn{1}{l|}{-2.04

                                          }               \\ \hline
\multicolumn{1}{|l|}{BDA}                     & -1

          &  +0.13

                      & -0.72

                                & \multicolumn{1}{l|}{-0.52

                                }               \\ \hline
\multicolumn{1}{|l|}{GFIR}                     & +2

          &  +0.13

                     &   -0.66

                             & \multicolumn{1}{l|}{+0.03
                             }               \\ \hline

\multicolumn{1}{|l|}{BMD}                     &  -5.25

         & +0.13

                      &   -0.59

                              & \multicolumn{1}{l|}{-1.08
                              }               \\ \hline
\multicolumn{1}{|l|}{SSR}                     &  -5.25

         &  +0.13

                     & +7.3

                                         & \multicolumn{1}{l|}{+3.38

                                      }               \\ \hline
\multicolumn{1}{|l|}{MSR}                     & -4.5

         &   +0.26

                    &  +4.21

                              & \multicolumn{1}{l|}{+1.78
                              }               \\ \hline
\multicolumn{1}{|l|}{ASSR}                    & -5.5

         &   +2.72

                    &  +2.24

                             & \multicolumn{1}{l|}{+1.23
                             }               \\ \hline
\multicolumn{1}{|l|}{LSSF}                    &   -5.75

       &   +0.39

                  &  -0.53

                         &
                          \multicolumn{1}{l|}{-1.04
                          }               \\ \hline
\multicolumn{1}{|l|}{Gradient}                    & -6.25
     &    0
      &   -4.99

             & \multicolumn{1}{l|}{-3.75
             }               \\ \hline
\multicolumn{1}{|l|}{PDF}                     & -0.25
  &     0
       &   -0.23

             & \multicolumn{1}{l|}{-0.58
             
             }               \\ \hline
\multicolumn{1}{|l|}{DCT}                &  -5
   &   0
       & +4.34

                & \multicolumn{1}{l|}{+1.71
                }               \\ \hline
\multicolumn{1}{|l|}{NPLE-QI}                    & -5
   &  +1.04

               &  +0.1

                       & \multicolumn{1}{l|}{-0.39
                       }               \\ \hline
\multicolumn{1}{|l|}{HOMO}                     &  -2.75
     &  0
       &  -1.64

                      & \multicolumn{1}{l|}{-1.34
                      }               \\ \hline
\multicolumn{1}{|l|}{NMBN}                 & -6

           &  +1.55

                       &   +0.53

                               & \multicolumn{1}{l|}{-0.15
                               }               \\ \hline
\end{tabular}
\end{table}

\noindent
Table \ref{Table2_1} shows the TP rates achieved by using the second manner of applying the pre-processing step. The GFIR obviously increase the TP rates in the ORL dataset. The second manner obviously improves the TP obtained by the NPLE-QI and PDF methods. There is small improvement obtained by applying the second method with the GFIR, SSR, MSR, ASSR, HOMO, and LSSF methods compared with the TR obtained by the first manner. However, most methods hurt the overall TP rate of the Viola-Jones algorithm except the PDF, HOMO, NPLE-QI, and GFIR methods. On the other hand, the FP rates are similar to what were achieved by the first manner ($\pm$ 2\%), as shown in Table \ref{Table2_2}.

\begin{table}[]
\centering
\caption{The effect of the pre-processing methods on the true positive rates (\%) using the manner 2.}
\label{Table2_1}
\begin{tabular}{l|l|l|l|l}
\cline{2-4}
                                              & \multicolumn{3}{c|}{\textbf{Datasets}}                    &                                     \\ \hline
\multicolumn{1}{|l|}{\textbf{Methods}} & ORL & MIT-CBCL & BioID &  \multicolumn{1}{l|}{\textbf{Total}} \\ \hline
\multicolumn{1}{|l|}{Sharpness}               & -24

           & -4.66

                       & -21.7

                                & \multicolumn{1}{l|}{-17.16

                                }               \\ \hline
\multicolumn{1}{|l|}{Wiener}                  &  -16.25

          &  -3.11

                      &   -80.6

                            & \multicolumn{1}{l|}{-48.83

                            }               \\ \hline
\multicolumn{1}{|l|}{BDA}                     & -11.75

         &   -1.036

                    &  -7.63
                          & \multicolumn{1}{l|}{-6.35

                                    }               \\ \hline
\multicolumn{1}{|l|}{GFIR}                     & \textbf{+3.5}

           &   -0.26

                     &  +0.72

                           & \multicolumn{1}{l|}{+0.85

                           }               \\ \hline

\multicolumn{1}{|l|}{BMD}                     &  -23.5

         &  -3.11

                    &    +0.59

                                   & \multicolumn{1}{l|}{-4.05

                                }               \\ \hline
\multicolumn{1}{|l|}{SSR}                     &  -11

          &  -1.55

                     &  -3.09

                            & \multicolumn{1}{l|}{-3.82

                            }               \\ \hline
\multicolumn{1}{|l|}{MSR}                     & -11

        &  -2.20

                  &   -3.16

                         & \multicolumn{1}{l|}{-4.05

                         }               \\ \hline
\multicolumn{1}{|l|}{ASSR}                    &  -27

      &    -7.38

               &    -30.51
                     & \multicolumn{1}{l|}{-23.36

                      }               \\ \hline
\multicolumn{1}{|l|}{LSSF}                    & -33

           &   -5.57

                    & -47.93

                             & \multicolumn{1}{l|}{-33.57

                             }               \\ \hline
\multicolumn{1}{|l|}{Gradient}                    & -78.5

          &   -85.75

                    &  -93.95

                           & \multicolumn{1}{l|}{-89.31

                           }               \\ \hline
\multicolumn{1}{|l|}{PDF}                     & 0

           &  +1.13

                       &  +3.18

                              & \multicolumn{1}{l|}{+2.12

                              }               \\ \hline
\multicolumn{1}{|l|}{DCT}                & -32.75

           &  -16.19

                      &   -2.7

                            & \multicolumn{1}{l|}{-11.03

                            }               \\ \hline
\multicolumn{1}{|l|}{NPLE-QI}                   &  -1.1
    & +1.31
     &   +2.51
        & \multicolumn{1}{l|}{+1.62

         }               \\ \hline
\multicolumn{1}{|l|}{HOMO}                     &  +1.3
     &  +1.2

              &   +2.5

                    & \multicolumn{1}{l|}{+1.94

                    }               \\ \hline
\multicolumn{1}{|l|}{NMBN}                 &  -27.25

          &  -5.96

                    &   -26.56

                          & \multicolumn{1}{l|}{-20.76

                          }               \\ \hline
\end{tabular}
\end{table}

\begin{table}[]
\centering
\caption{The effect of the pre-processing methods on the false positive rates (\%) using the manner 2.}
\label{Table2_2}
\begin{tabular}{l|l|l|l|l}
\cline{2-4}
                                              & \multicolumn{3}{c|}{\textbf{Datasets}}                    &                                     \\ \hline
\multicolumn{1}{|l|}{\textbf{Methods}} & ORL & MIT-CBCL & BioID &  \multicolumn{1}{l|}{\textbf{Total}} \\ \hline
\multicolumn{1}{|l|}{Sharpness}              &  -4.75
      &  +2.07

                  &   -0.72

                          & \multicolumn{1}{l|}{-0.52

                          }               \\ \hline
\multicolumn{1}{|l|}{Wiener}                  & -5.25

          &   +2.46

                    &   -3.68

                           & \multicolumn{1}{l|}{-2.15

                           }               \\ \hline
\multicolumn{1}{|l|}{BDA}                     &  -0.75

          &  +0.13

                      &    -0.72

                            & \multicolumn{1}{l|}{-0.48
                            }               \\ \hline
\multicolumn{1}{|l|}{GFIR}                     &  +1.75

         &   +0.13

                  &  -0.53

                          & \multicolumn{1}{l|}{-0.002

                          }               \\ \hline

\multicolumn{1}{|l|}{BMD}                     & -4.75

          &  +0.13

                     & +0.66

                              & \multicolumn{1}{l|}{-0.30
                              }               \\ \hline
\multicolumn{1}{|l|}{SSR}                     & -5.25

           &   +0.13

                     &  +4.73

                             & \multicolumn{1}{l|}{+1.93
                             }               \\ \hline
\multicolumn{1}{|l|}{MSR}                     & -4.75

           & +0.26

                        &  +3.02

                                & \multicolumn{1}{l|}{+1.08
                                }               \\ \hline
\multicolumn{1}{|l|}{ASSR}                    & -5.25

           &   +1.68

                     & +2.37

                              & \multicolumn{1}{l|}{+1.04
                              }               \\ \hline
\multicolumn{1}{|l|}{LSSF}                    & -6

         &  +0.39

                     &  -0.46

                              & \multicolumn{1}{l|}{-1.04
                              }               \\ \hline
\multicolumn{1}{|l|}{Gradient}                    & -6.25
   &   0
        & -5.13

                  & \multicolumn{1}{l|}{-3.82
                  }               \\ \hline
\multicolumn{1}{|l|}{PDF}                     & -2.75
    &    +0.13

              & +0.2

                       & \multicolumn{1}{l|}{-0.28
                       
                       }               \\ \hline
\multicolumn{1}{|l|}{DCT}                & -4.75

          &   0

                    &  +2.56

                            & \multicolumn{1}{l|}{+0.74
                            }               \\ \hline
\multicolumn{1}{|l|}{NPLE-QI}                    &  -4.75
  &  +1.16

               &   -3.89

                               & \multicolumn{1}{l|}{-2.57

                            }               \\ \hline
\multicolumn{1}{|l|}{HOMO}                     &  -3.5
    &  0
      & -0.92
            & \multicolumn{1}{l|}{-1.04

                       }               \\ \hline
\multicolumn{1}{|l|}{NMBN}                 & -5.75

          &   +1.43

                   &   +1.05

                         & \multicolumn{1}{l|}{+0.15

                         }               \\ \hline
\end{tabular}
\end{table}
\noindent
 As expected, the number of the undetected faces using the first and  second manners is increased for most methods,  as shown in Table \ref{Table1_3} and Table \ref{Table2_3}. The experimental results show that both manners have a similar false negative (FN) rate using all of the used pre-processing methods, except some methods that are improved using the second manner.
 \\
 Eventually, the PDF is considered the best pre-processing method that increases the detection accuracy by 0.41\%, 2.12\% using the first, and second manners, respectively. The HOMO and the NPLE-QI are the second top methods that improve the TP rate by 1.94\% and 1.62\%, respectively, using the second method. However, the improvements obtained by both methods and the GFIR method using the first manner are considered very small. There is a common factor among the best methods; the generated image does not have a significant difference in intensity from the original image. That is expected, as the Viola-Jones face detector is based on the similar natural properties of human faces that are destroyed by some pre-processing methods, e.g. gradient normalization. As shown in Fig. 4, pixel intensities are dramatically changed within the pre-processing step, e.g. sharpness, and Wiener, ASSR, and NMBN, which may mislead the Viola- Jones face detector instead of boosting it.
 \\
 To that end, we adopt the GFIR, the NPLE-QI, the HOMO and the PDF methods as the best pre-processing methods that have a potential impact on increasing the accuracy of the Viola-Jones algorithm and the smallest chance 
 of hurting it. However, this experiment does not show us how much improvement is possible using these pre-processing methods when the face images suffer from bad conditions.


\begin{figure}
\centering
\includegraphics[width=0.9\linewidth]{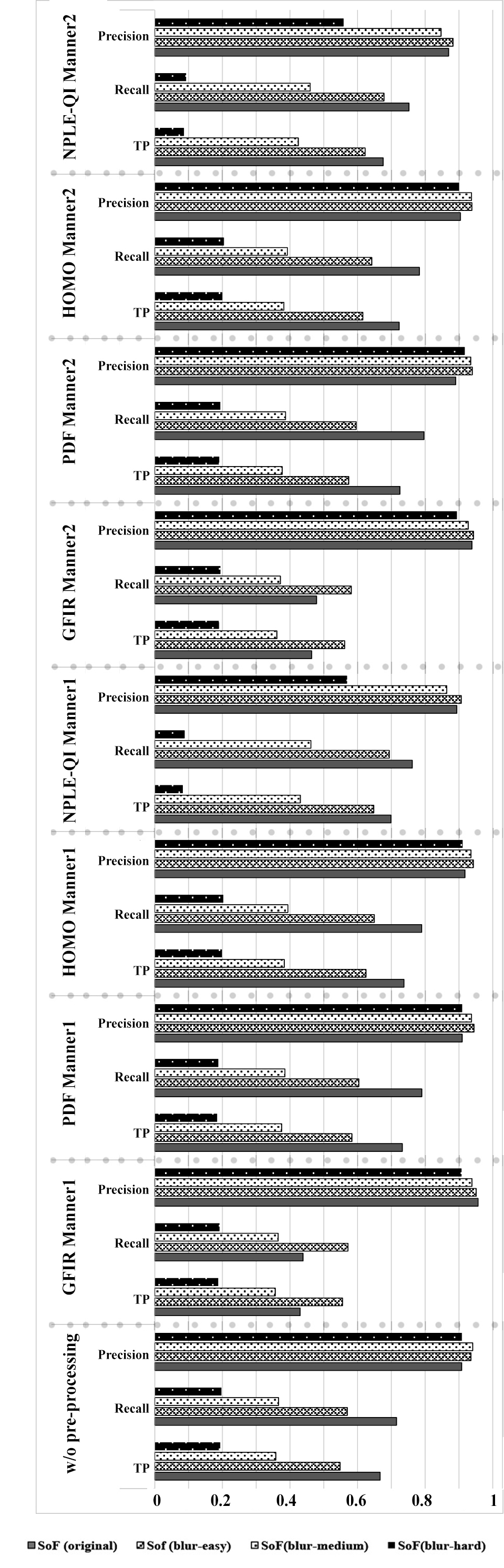}
\caption{The TP rate, recall, and precision obtained by the Viola-Jones algorithm with/without the four pre-processing methods using the SoF dataset. }
\end{figure}


\begin{table}[]
\centering
\caption{The effect of the pre-processing methods on the false negative rates (\%) using the manner 1.}
\label{Table1_3}
\begin{tabular}{l|l|l|l|l}
\cline{2-4}
                                              & \multicolumn{3}{c|}{\textbf{Datasets}}                    &                                     \\ \hline
\multicolumn{1}{|l|}{\textbf{Methods}} & ORL & MIT-CBCL & BioID &  \multicolumn{1}{l|}{\textbf{Total}} \\ \hline
\multicolumn{1}{|l|}{Sharpness}             &   +29

        &  +2.59

                    &  +20.05

                            & \multicolumn{1}{l|}{+16.38
                            }               \\ \hline
\multicolumn{1}{|l|}{Wiener}                  & +20.75

        &   +0.65

                   &  +80.8

                            & \multicolumn{1}{l|}{+48.90
                            }               \\ \hline
\multicolumn{1}{|l|}{BDA}                     & +12.5

       &  +0.91

                  &    +5.79

                        & \multicolumn{1}{l|}{+5.38
                        }               \\ \hline
\multicolumn{1}{|l|}{GFIR}                     & -2.25

          &  +0.26

                     &  -0.46
                            & \multicolumn{1}{l|}{-0.52
                            }               \\ \hline

\multicolumn{1}{|l|}{BMD}                     & +29.75
    &  +2.98
      	
                 & -0.46

                           & \multicolumn{1}{l|}{+5.01
                           }               \\ \hline
\multicolumn{1}{|l|}{SSR}                     &  +17.25

        &  +1.42

                  &  +2.76

                          & \multicolumn{1}{l|}{+4.53
                          }               \\ \hline
\multicolumn{1}{|l|}{MSR}                     & +15

         &   +1.94

                   &  +4.14

                         & \multicolumn{1}{l|}{+5.12
                         }               \\ \hline
\multicolumn{1}{|l|}{ASSR}                    &  +30

        & +5.83

                   &  +28

                            & \multicolumn{1}{l|}{+21.95
                            }               \\ \hline
\multicolumn{1}{|l|}{LSSF}                    &  +39.5

         &  +5.57

                     &  +48.13

                             & \multicolumn{1}{l|}{+34.65
                             }               \\ \hline
\multicolumn{1}{|l|}{Gradient}                    & +84.75

           &  +85.75

                       & +95.6

                                & \multicolumn{1}{l|}{+91.16
                                }               \\ \hline
\multicolumn{1}{|l|}{PDF}                     &  +3.25

          & -0.52

                      &    -0.2

                            & \multicolumn{1}{l|}{+0.29
                            
                            }               \\ \hline
\multicolumn{1}{|l|}{DCT}                &  +37

          & +16.84

                      & +3.09

                               & \multicolumn{1}{l|}{+12.07
                               }               \\ \hline
\multicolumn{1}{|l|}{NPLE-QI}                    & +17
      &  +5.96

                &   +1.12

                       & \multicolumn{1}{l|}{+4.86
                       }               \\ \hline
\multicolumn{1}{|l|}{HOMO}                     & +2.5
    &   -0.39

               &   +1.4
                    & \multicolumn{1}{l|}{+1.1
                    }               \\ \hline
\multicolumn{1}{|l|}{NMBN}                 &  +34.5

          &   +4.92

                    &  +25.12

                            & \multicolumn{1}{l|}{+20.72

                            }               \\ \hline
\end{tabular}
\end{table}

\begin{table}[]
\centering
\caption{The effect of the pre-processing methods on the false negative rates (\%) using the manner 2.}
\label{Table2_3}
\begin{tabular}{l|l|l|l|l}
\cline{2-4}
                                              & \multicolumn{3}{c|}{\textbf{Datasets}}                    &                                     \\ \hline
\multicolumn{1}{|l|}{\textbf{Methods}} & ORL & MIT-CBCL & BioID &  \multicolumn{1}{l|}{\textbf{Total}} \\ \hline
\multicolumn{1}{|l|}{Sharpness}             &  +28.75

          & +2.59

                       & +20.05

                                & \multicolumn{1}{l|}{+16.34
                                }               \\ \hline
\multicolumn{1}{|l|}{Wiener}                  & +21.5

           &  +0.65

                       &    80.8

                              & \multicolumn{1}{l|}{+49.02

                              }               \\ \hline
\multicolumn{1}{|l|}{BDA}                     &   +12.5

        & +0.91

                     & +5.85
                           & \multicolumn{1}{l|}{+5.42

                                     }               \\ \hline
\multicolumn{1}{|l|}{GFIR}                     & -2.25
     & +0.13

                 &  -0.46
                      & \multicolumn{1}{l|}{-0.56

                      }               \\ \hline

\multicolumn{1}{|l|}{BMD}                     &  +28.5
    &    +2.98

            &  -0.33

                             & \multicolumn{1}{l|}{+4.90
                             }               \\ \hline
\multicolumn{1}{|l|}{SSR}                     & +16.25

          &  +1.42

                      &   +3.02

                              & \multicolumn{1}{l|}{+4.53
                              }               \\ \hline
\multicolumn{1}{|l|}{MSR}                     &  +15.75

        &  +1.81

                    & +3.02

                              & \multicolumn{1}{l|}{+4.57
                              }               \\ \hline
\multicolumn{1}{|l|}{ASSR}                    & +32.25

           &  +5.7

                       &  +26.96
                              & \multicolumn{1}{l|}{+21.65

                                       }               \\ \hline
\multicolumn{1}{|l|}{LSSF}                    &  +39

          &  +5.18

                    &   +46.94

                            & \multicolumn{1}{l|}{+33.79
                            }               \\ \hline
\multicolumn{1}{|l|}{Gradient}                    & +84.75

            &   +85.1

                      & +95.6

                              & \multicolumn{1}{l|}{+90.98
                              }               \\ \hline
\multicolumn{1}{|l|}{PDF}                     & +3

          &  -1.26

                      &   -3.2

                              & \multicolumn{1}{l|}{-1.72
                              }               \\ \hline
\multicolumn{1}{|l|}{DCT}                & +37.5

          &    +16.2

                    &   +3.22

                            & \multicolumn{1}{l|}{+12.03
                            }               \\ \hline
\multicolumn{1}{|l|}{NPLE-QI}                    &  +5.85
    &   +0.15

             &  +1.38

                     & \multicolumn{1}{l|}{+1.69

                     }               \\ \hline
\multicolumn{1}{|l|}{HOMO}                     &  +2.2

          &  -1.2

                    &   +3.42
                         & \multicolumn{1}{l|}{+1.91
                         }               \\ \hline
\multicolumn{1}{|l|}{NMBN}                 &    +33

     &    +4.79

              &  +24.79
                   & \multicolumn{1}{l|}{+20.27
                   }               \\ \hline
\end{tabular}
\end{table}

\subsection{Second set of experiments}
In this set of experiments, we have used only the pre-processing methods that either achieved some improvement in the first set of experiments or had the lowest probability of hurting the face detector's accuracy, namely the GFIR, the HOMO, the PDF, and the NPLE-QI methods. We have used the SoF dataset that contains many low-light, occluded, non-frontal, and blurry face images. We have applied the two manners that have been discussed previously. The original Viola-Jones algorithm has obtained 66.68\%, 54.83\%, 35.81\%, and 19.32\% TP rates using the original set of images and the easy, the medium, and the hard sets of blurry face images, respectively. The FP rates obtained by the Viola-Jones are 6.76\%, 3.76\%, 2.25\%, and 1.95\% using the aforementioned sets, respectively. Eventually, the FN rates were 26.56\%, 41.41\%, 61.93\%, and 78.73\%, respectively. As shown in Fig. 8, the TP rate is improved by all pre-processing methods except the GFIR method that hurts the TP rate using all sets of images except the medium set, using the second manner, and the easy and medium sets, using both manners. In the original set of the SoF dataset, the HOMO and the PDF methods achieve the best TP rate using both manners. The HOMO method improves the TP rate by 7.06\% and 5.63\% using the first and second manners, respectively. The PDF method improves the TP rate by 6.57\% and 5.86\% using the first and second manners, respectively. The NPLE-QI method achieves the best improvement of the TP rate using the easy set of blurry images by increasing the TP rate of the Viola-Jones face detector by 9.96\% and 7.4\% using the first and second manners, respectively. Additionally, it gets the best TP rate in the medium set of the blurry faces by improving the original TP rate by 7.25\% and 6.25\% using the first and second manners, respectively. From another viewpoint, the recall is improved by all methods except the GFIR method. The only set that has some improvement obtained by the GFIR is the easy set of blurry faces, where, it improves the recall rate by around 1\%. The PDF and HOMO methods achieve the best improvement. The PDF method increases the recall by 7.5\% and 8.15\% using the first and second manner, respectively. The HOMO method improves the rate by 7.45\% and 6.8\% using the first and second manner, respectively. As the NPLE-QI obtains the best TP rate in the easy and medium sets of blurry faces, it obtains the best improvement, in term of recall, in the same sets using both manners. In term of precision, the NPLE-QI method hurts the precision rate using the hard level set of blurry face images. The obvious improvement obtained by GFIR is in the precision rate using the original set of images, where, it improves the precision rate by around 4\%.

\subsubsection{Analysis}

Tables 7 and 8 illustrate the improvement of TP, FP, and FN rates obtained by the four pre-processing methods using the four sets of images, namely the original images of the SoF dataset and the three sets of blurry face images. As shown, the GFIR method does not have an obvious improvement; indeed it hurts the face detector using the original set of the SoF dataset. As expected, the GFIR increases the FN rate using both manners. All other methods have an obvious improvement in terms of TP, FP, and FN rates. However, all methods either hurt or have no evident improvement using the hard set of blurry face images of the SoF dataset. Fig. 9 illustrates the difficulty of each set of the blurry faces provided in the SoF dataset. As shown, the hard level is considered an extreme case of difficulty; for that reason, the pre-processing methods did not achieve any improvement, except a small level of improvement by the HOMO method. Although the GFIR is a deblurring method, it hurt the face detector using the second and third set of blurry face images. This is due to the fixed parameters we use in the blind pre-processing stage.
\\
As a sanity check, we have tested another pre-trained Cascade classifier, namely the OpenCV pre-trained Cascade classifier of the Viola-Jones face detector. Fig. 10 shows the TP rates obtained by both the Matlab/OpenCV Viola-Jones face detector using the four sets of the SoF dataset, namely the original images and the three sets of blurry face images. As shown, the pre- processing methods almost have the same effect on improving the TP rates obtained by the Viola-Jones face detector, except the NPLE-QI using the second manner.

\subsection{Time analysis}
As shown above, the GFIR method is not robust enough in term of improving the Viola-Jones face detector. Thus, we can exclude it. We have studied the time required by the three suggested pre-processing methods, namely the HOMO, the PDF, and the NPLE-QI methods. The HOMO method takes
35.86\% and 44.96\% of the time required by the Matlab and OpenCV implementations of the Viola-Jones face detector, respectively. The PDF method takes 33.95\% and 42.57\% of the time required by the Matlab and OpenCV implementations of the Viola-Jones face detector, respectively. Eventually, the NPLE-QI takes 852.57\% and 1068.86\% of the time required by the Matlab and OpenCV implementations of the Viola-Jones face detector, respectively. 
As shown, from both performance and accuracy viewpoints, the PDF and HOMO methods are considered the best pre-processing method, followed by the NPLE-QI method.
\begin{figure*}
\centering
\includegraphics[width=0.8\linewidth]{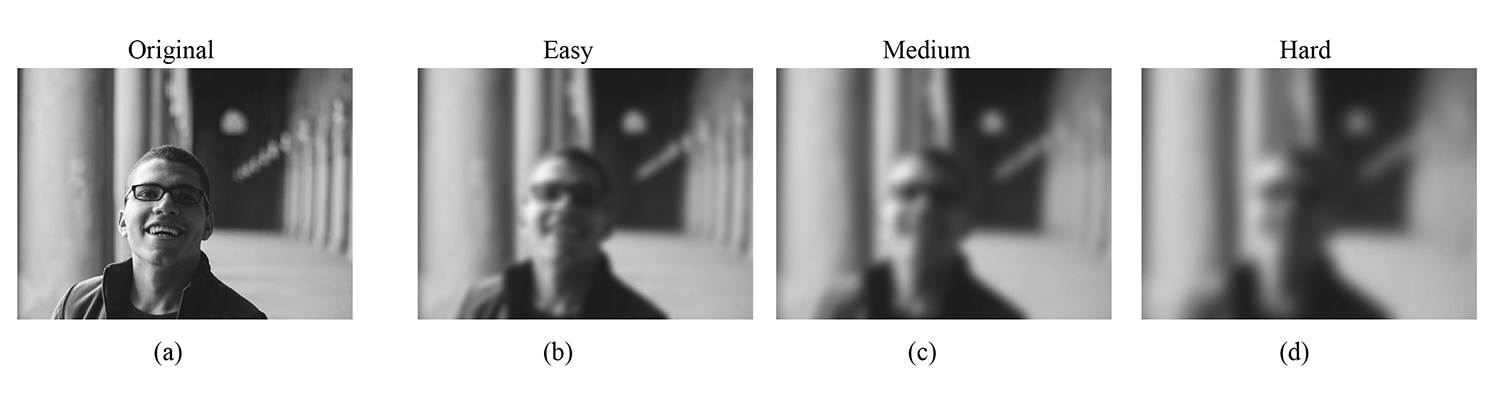}
\caption{The difficulty of each group of blurry faces in the SoF dataset. (a) Original image. (b) Easy level. (c) Medium level. (d) Hard level. }
\end{figure*}

\begin{table*}[]
\centering
\caption{The effect of the four pre-processing methods on the TP, FP, and FN rates (\%) using the first manner.}

\begin{tabular}{|l|l|l|l|l|l|}
\hline
\multicolumn{2}{|c|}{\textbf{Method}} & \multicolumn{1}{c|}{\textbf{Orignal}} & \multicolumn{1}{c|}{\textbf{Easy}} & \multicolumn{1}{c|}{\textbf{Medium}} & \multicolumn{1}{c|}{\textbf{Hard}} \\ \hline
\multirow{3}{*}{GFIR} & TP & -23.67 & +0.71 & -0.15 & -0.56 \\ \cline{2-6} 
 & FP & -4.81 & -0.90 & +0.08 & -0.04 \\ \cline{2-6} 
 & FN & +28.47 & +0.19 & +0.08 & +0.60 \\ \hline
\multirow{3}{*}{PDF} & TP & \textbf{+6.57} & +3.49 & +1.73 & -0.90 \\ \cline{2-6} 
 & FP & +0.53 & -0.34 & +0.26 & -0.11 \\ \cline{2-6} 
 & FN & \textbf{-7.10} & -3.16 & -1.99 & +1.01 \\ \hline
\multirow{3}{*}{HOMO} & TP & \textbf{+7.06} & \textbf{+7.67} & +2.52 & +0.53 \\ \cline{2-6} 
 & FP & -0.15 & +0.04 & +0.38 & 0.00 \\ \cline{2-6} 
 & FN & \textbf{-6.91} & \textbf{-7.70} & -2.89 & -0.53 \\ \hline
\multirow{3}{*}{NPLE-QI} & TP & +3.19 & \textbf{+9.96} &\textbf{ +7.25} & -11.05 \\ \cline{2-6} 
 & FP & +1.54 & +2.89 & +4.51 & +4.32 \\ \cline{2-6} 
 & FN & -4.73 & \textbf{-12.85} & \textbf{-11.76} & +6.73 \\ \hline
\end{tabular}
\end{table*}

\begin{table*}[]
\centering
\caption{The effect of the four pre-processing methods on the TP, FP, and FN rates (\%) using the second manner.}

\begin{tabular}{|l|l|l|l|l|l|}
\hline
\multicolumn{2}{|c|}{\textbf{Method}} & \multicolumn{1}{c|}{\textbf{Orignal}} & \multicolumn{1}{c|}{\textbf{Easy}} & \multicolumn{1}{c|}{\textbf{Medium}} & \multicolumn{1}{c|}{\textbf{Hard}} \\ \hline
\multirow{3}{*}{GFIR} & TP & -20.29 & +1.35 & +0.30 & -0.34 \\ \cline{2-6} 
 & FP & -3.72 & -0.38 & +0.56 & +0.30 \\ \cline{2-6} 
 & FN & +24.00 & -0.98 & -0.86 & +0.04 \\ \hline
\multirow{3}{*}{PDF} & TP & \textbf{+5.86} & +2.56 & +1.88 & -0.30 \\ \cline{2-6} 
 & FP & +2.18 & -0.07 & +0.38 & -0.23 \\ \cline{2-6} 
 & FN & \textbf{-8.04} & -2.49 & -2.25 & +0.53 \\ \hline
\multirow{3}{*}{HOMO} & TP & \textbf{+5.63} & \textbf{+6.76} & +2.41 & +0.64 \\ \cline{2-6} 
 & FP & +0.90 & +0.30 & +0.30 & +0.26 \\ \cline{2-6} 
 & FN & \textbf{-6.54} & \textbf{-7.07} & -2.71 & -0.90 \\ \hline
\multirow{3}{*}{NPLE-QI} & TP & +0.90 & \textbf{+7.40} & \textbf{+6.65} & -10.71 \\ \cline{2-6} 
 & FP & +3.38 & +4.51 & +5.41 & +4.85 \\ \cline{2-6} 
 & FN & -4.28 & \textbf{-11.91} & \textbf{-12.06} & +5.86 \\ \hline
\end{tabular}
\end{table*}
\begin{figure}
\centering
\includegraphics[width=0.9\linewidth]{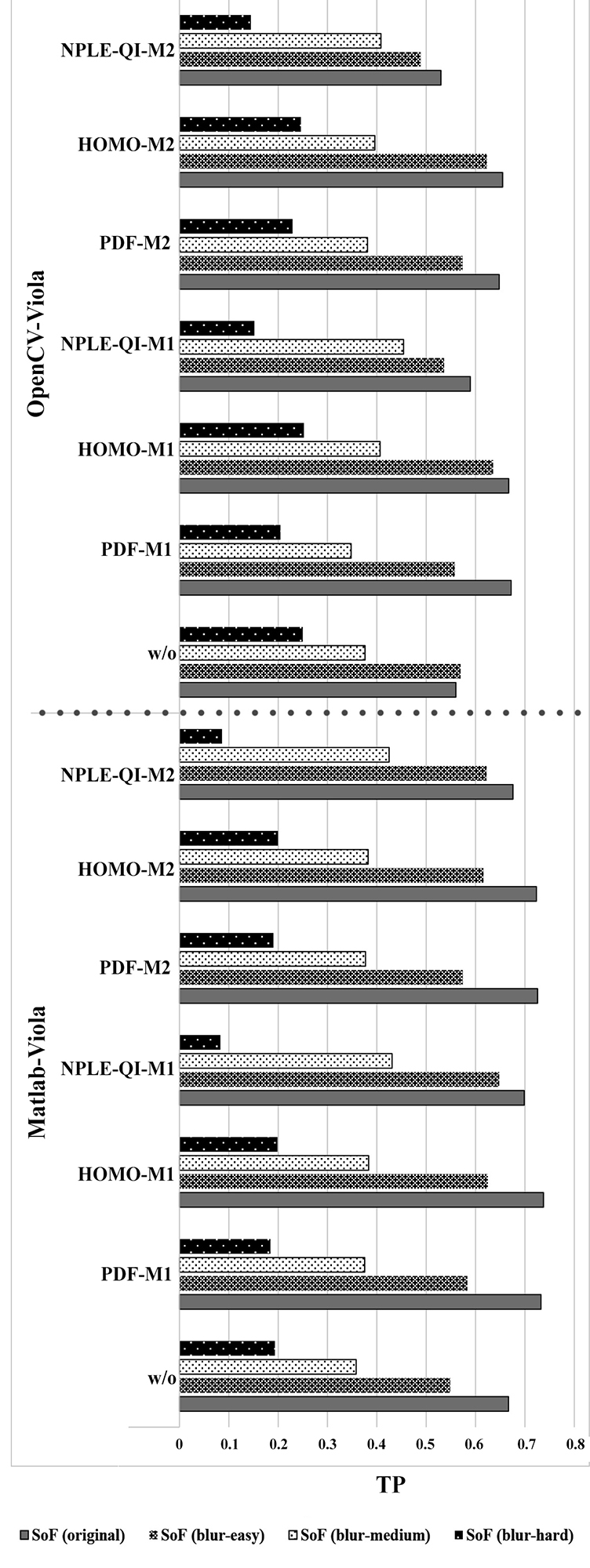}
\caption{The TP rate obtained by the Viola-Jones face detector (Matlab/OpenCV). }
\end{figure}

\section{Conclusions}
We have presented a study that aims to improve the detection accuracy of the pre-trained Viola-Jones face detector using a set of pre-processing methods instead of using complicated detection approaches. In the pre-processing stage, two main categories of image enhancement methods have been applied. The first category focuses on performing a photometric normalization to get lighting invariant images that enhance dark images or badly illuminated faces. The second category consists of deblurring methods that reconstruct blurred images. There are two strategies of applying the pre-processing methods. The first manner is a blind pre-processing stage that is performed for all images before the face detection process. The second manner adds an image adjustment module before the face detection stage. 
\\
In order to draw a coherent conclusion about the potential improvement caused by the pre-processing methods, we have performed two sets of experiments. In the first set, ten lighting pre-processing methods and five deblurring and sharpening methods have been applied to 2,693 face images obtained from three different datasets. The goal of this set of experiments was to discard any pre-processing method that may hurt the Viola-Jones face detector. The experimental results show that all the pre-processing methods hurt the detection rate of the Viola-Jones algorithm, except one deblurring method and three photometric normalization methods. In the second set of experiments, these four methods, the GFIR, the HOMO, the NPLE-QI, and the PDF methods, have been used using a hard face dataset that suffers from many bad lighting conditions and has many blurry face images. The experimental results show that the deblurring method, i.e. the GFIR method, does not improve the true positive (TP) rate of the Viola-Jones algorithm. However, there is an obvious improvement on the TR rate when we use the photometric normalization methods, namely the PDF, the HOMO, and the NPLE-QI methods, using both manners.
\\
Overall, we found that by using some simple and fast blind photometric normalization methods, namely PDF, HOMO and NPLE-QI, as a pre-processing step, the accuracy of the Viola-Jones face detector has been obviously improved with a small chance of hurting the pre-trained Cascade classifier. This encourages people who use the ready-to-use Viola-Jones face detector in vision-based applications to use these methods in order to improve the face detection accuracy.

\end{document}